# Machine Learning for Missing Value Imputation

Abu Fuad Ahmad, Khaznah Alshammari, Istiaque Ahmed, and MD Shohel Sayed

*Abstract*—In recent times, a considerable number of research studies have been carried out to address the issue of Missing Value Imputation (MVI). MVI aims to provide a primary solution for datasets that have one or more missing attribute values. The advancements in Artificial Intelligence (AI) drive the development of new and improved machine learning (ML) algorithms and methods. The advancements in ML have opened up significant opportunities for effectively imputing these missing values. The main objective of this article is to conduct a comprehensive and rigorous review, as well as analysis, of the state-of-the-art ML applications in MVI methods. This analysis seeks to enhance researchers' understanding of the subject and facilitate the development of robust and impactful interventions in data preprocessing for Data Analytics. The review is performed following the Preferred Reporting Items for Systematic Reviews and Meta-Analysis (PRISMA) technique. More than 100 articles published between 2014 and 2023 are critically reviewed, considering the methods and findings. Furthermore, the latest literature is examined to scrutinize the trends in MVI methods and their evaluation. The accomplishments and limitations of the existing literature are discussed in detail. The survey concludes by identifying the current gaps in research and providing suggestions for future research directions and emerging trends in related fields of interest.

*Impact Statement* — **This is a comprehensive review that explores the application of machine learning techniques in handling missing data across various domains. The study examines existing literature and methodologies, highlighting the significance of ML-based approaches in addressing the challenge of missing values in datasets. The paper elucidates the strengths and limitations of different ML algorithms and their effectiveness in imputing missing values across different types of data. Additionally, the research identifies emerging trends, challenges, and future research directions in the field of missing value imputation using ML. The findings of this research contribute to advancing the understanding of ML techniques for handling missing data and provide valuable insights for practitioners and researchers seeking to develop more robust and accurate imputation methods in data analysis and machine learning applications.**

## I. INTRODUCTION

ARTIFICIAL Intelligence (AI) and Machine Learning (ML) are two interconnected fields. AI encompasses the development of intelligent systems that can perform tasks that typically require human intelligence. ML is a key component of many AI systems as it provides algorithms and methods for learning from data. As ML techniques improve and evolve, they enable AI systems to become more capable, efficient, and accurate in performing a wide range of tasks. ML techniques serve as the foundation for many AI applications and advancements in AI drive the development of new and improved ML algorithms and methods.

The term "machine learning" is commonly used in various fields, but there is no universally agreed-upon definition. This lack of consensus can be attributed to the broad range of areas it encompasses and the diverse contributions of researchers from different disciplines who have been involved in its development. In a broad sense, machine learning refers to an algorithmic framework that enables the analysis of data, facilitates inference, and offers a preliminary approach to establishing functional relationships [1].

Machine learning (ML) comprises a class of data science models that have the ability to learn from data and enhance their performance over time. The roots of ML can be traced back to the scientific community's interest in the 1950s and 1960s in replicating human learning through computer programs. From this perspective, ML extracts knowledge from data, which can then be utilized for prediction and generating new information. This knowledge helps to reduce uncertainty by providing guidance on how to solve specific problems. ML is particularly valuable when dealing with tasks that do not have explicit instructions for an analytic solution, such as image and voice processing, pattern recognition, and complex classification tasks. ML models have gained popularity in both academic research and industry due to their superior performance in

¹This paragraph of the first footnote will contain the date on which you submitted your paper for review. It will also contain support information, including sponsor and financial support acknowledgment.

Abu Fuad Ahmad is with the department of Computer Science of New Mexico State University, Las Cruces, NM 88001 USA (e-mail: fuad@nmsu.edu).

Khaznah Alshammari is with the department of Computer Science of New Mexico State University, Las Cruces, NM 88001 USA, on leave from Northern Border University, Rafha, Saudi Arabia (e-mail: Kalshamm@nmsu.edu).

Istiaque Ahmed is with the Graduate School of Informatics, Osaka Metropolitan University, Osaka 558-8585, Japan (e-mail: sw23837u@st.omu.ac.jp).

MD Shohel Sayed is with the Multimedia University, Melaka 75450, Malaysia (e-mail: shohel.sayeed@mmu.edu.my).

Associate Editor: MD Shohel Sayed (e-mail: shohel.sayeed@mmu.edu.my).

processing, classifying, and forecasting using complex and large-scale data [2].

A dataset serves as the core component of any automated classification or regression system that aims to make learnable decisions. However, practical datasets collected in real-world scenarios often exhibit an unusual proportion of missing values, irregular patterns (outliers), and redundancy across attributes [3]. These issues need to be addressed and resolved in order to develop a robust and generalizable trained model. Traditionally, missing values are represented as NaNs, blanks, undefined, null, or other placeholders [4], [5]. There are multiple reasons for such missingness, stemming from various sources within the datasets. These reasons include incorrect or erroneous data entries, unavailability of data, data collection problems, missing sequences, incomplete features, missing files, incomplete information, and more. Regardless of the specific causes, it is crucial to handle missing data because relying on statistical results derived from datasets with missing values can introduce biases. Additionally, missing data introduces unavoidable ambiguity in statistical analyses. Furthermore, it is worth noting that many Machine Learning (ML) algorithms do not support datasets with missing values [5], [6].

Various patterns of missing values are often exhibited by data, each possessing distinct characteristics and implications for data analysis. The understanding of these patterns is deemed essential for the selection of appropriate imputation techniques and the interpretation of results. Missing data is categorized into three fundamental types: MCAR (Missing Completely At Random), MAR (Missing At Random), and MNAR (Missing Not At Random) [7], [8]. In MCAR scenarios, the absence of data points is entirely random and not related to any observed or unobserved variables. The probability of data being missing does not depend on the values of other data points. MAR refers to cases where the probability of missing data depends on observed variables but not on unobserved data. In other words, the missingness is related to the available data but not to the data itself. This type of missingness is more complex than MCAR and requires careful handling. MNAR, sometimes called non-ignorable missingness, occurs when the probability of data being missing depends on unobserved data or factors not included in the dataset. MNAR data are the most challenging to handle, as they introduce potential bias and complicate imputation [6], [9], [10].

We are not the first researchers to delve into reviewing the applications of Machine Learning in missing value imputation. There have been several analyses, surveys and papers in related or adjacent scientific fields that have explored ML methodologies in different contexts [11], [12], [13], [14], [15], [16], [17], [18]. Suresh et al. [19] assessed the efficacy of data imputation methods utilizing 12 machine learning algorithms, spanning from linear regression to neural networks. The findings indicated that ensemble algorithms consistently exhibited superior imputation performance compared to the benchmark linear regression. In a separate comparative study conducted by Kokla et al. [20], nine imputation techniques for handling missing metabolic data were evaluated. The experiments revealed that, for both MAR and MCAR scenarios, random forest (RF) yielded the lowest normalized RMSE in the estimation of missing values. Tsai et al. [21] conducted a comparative study on five widely recognized supervised learning techniques: KNN, MLP, CART, naive Bayes, and SVM. They assessed the imputation results for categorical, numerical, and mixed data types. The findings revealed that CART outperforms other methods in handling categorical datasets, while MLP demonstrates optimal performance for numerical and mixed datasets in terms of classification accuracy. However, Kamrul et al. [22] presented a comprehensive review and analysis of statistics-based and ML-based MVI methods published from 2010 to 2021 using the PRISMA technique.

| LotArea | LotFrontage | MasVnrType | MasVnrArea | BsmtQual | GarageType |
|---|---|---|---|---|---|
| 21453 | NaN | NaN | 0.0 | TA | Attchd |
| 5604 | 67.0 | NaN | 0.0 | TA | NaN |
| 7301 | 64.0 | BrkFace | 500.0 | NaN | BuiltIn |
| 12692 | NaN | NaN | 0.0 | Gd | Attchd |
| 2117 | NaN | BrkFace | 513.0 | Gd | Detchd |
| 8963 | NaN | BrkFace | 289.0 | NaN | BuiltIn |
| 7000 | NaN | BrkFace | 90.0 | TA | Attchd |

Fig. 1. Partial dataset from the Housing data containing missing values represented as NaN.

Lin and Tsai [23] conducted an evaluation of statistical analyses pertaining to technical questions related to experimental design, such as the missing ratio, domain, and missing mechanism of experimental datasets used in various literature from 2006 to 2017. Another review study [24] approached the topic from different dimensions, including methodologies, experimental setups, and evaluation metrics used in existing data imputation techniques, particularly in the ML field from 2010 to 2020. The study revealed that clustering- and instance-based algorithms are the most proposed MVI methods, with the Percentage of Correct Prediction (PCP) and RMSE being the most commonly used evaluation metrics. Reference [25] critically reviews state-of-the-art practices found in the literature for addressing missing data in both Machine Learning and Deep Learning algorithms. In a related effort, Reference [6] delves into various methods for imputing missing data specifically in time-series data, offering a prototype outlining essential steps for constructing a model dedicated to imputing missing values in time-stamped data.

These existing survey papers and reviews contribute valuable insights and perspectives to the broader understanding of ML applications in related fields. Our study complements and extends their findings, with a specific emphasis on the applications of ML in imputation methods. It thoroughly examines the experimental setups, dataset characteristics, and evaluation metrics employed by innovative methods proposed during the period spanning 2014 to 2023. This study demonstrates that the appropriate application of MVI methods with ML models effectively enhances the performance of decision-making actions. We synthesized outcomes from numerous articles that employed different MVI and ML models for various datasets in different domains. Our review paper focuses on addressing following questions:

1) *Which MVI methods have been frequently used and which ones are underutilized, along with the reasons behind this discrepancy?*
2) *What are the popular application domains of MVI methods, and which domains remain under-explored?*
3) *What are the potential opportunities for future research in these areas?*

The core contributions of this review article on MVI methods provide support to the research community by addressing the aforementioned questions and offering a compilation of ten years' worth of different MVI methods for missingness imputation (2014-2023). This compilation includes the recent trends of commonly employed techniques and their evaluation. The article also presents sufficient evidence to prove that incorporating MVI methods can enhance the performance of ML models. It showcases results from diverse articles that cover various approaches and domain datasets. Additionally, the review provides satisfactory descriptions of MVI methods and ML models, along with their respective evaluations. Furthermore, it offers several recommendations for selecting these methods and models to develop comprehensive decision-making systems for real-world applications in diverse domains.

The remaining sections of the paper are structured as follows: Section II provides the detailed systematic search methodology. In Section III, a compilation of different ML methods from the selected articles is presented. Section IV and section V provide an overview of MVI methods, including a detailed explanation of their technical aspects, categorization, evaluation metrics, systematic review, and trends observed in their utilization over the years. Section VI focuses on a relative comparison of the discussed methods and offers comprehensive illustrations of recent trends in their usage. Furthermore, this section includes a thorough discussion and analysis of various computational pipelines that combine ML techniques and MVI methods across diverse datasets in different application domains. It highlights the impact of MVI strategies on the decision-making outcomes of ML models. Then, in section VII, several noteworthy observations and recommendations are provided. Final section summarizes the conclusions derived from the study.

## II. METHODOLOGY

### A. Systematic searches and data sources

The systematic review followed the guidelines of the Preferred Reporting Items for Systematic Reviews and Meta-Analyses (PRISMA) [26]. PRISMA is a comprehensive guideline designed to ensure transparent and complete reporting in systematic reviews and meta-analyses. All retrieved records were imported into Mendeley, a bibliography management software provided by Mendeley Ltd [27]. Using the keywords "Missing Value Imputation," a search was conducted in the Google Scholar databases within the title, abstract, and keywords of the articles. The search specifically focused on studies published after 2013 to include the most up-to-date evidence. To minimize selection bias, articles are independently screened by the authors and resolved any disagreements through discussion and consensus. The screening process involved reviewing the abstracts for inclusion and reading the full text of eligible articles based on the predefined criteria. Articles that lacked sufficient information on the application of machine learning or did not meet the eligibility criteria were excluded. The search syntax for ML techniques was developed based on relevant literature contexts [11], [12], [22] [23]. Additionally, the search was limited to studies published in the English language.

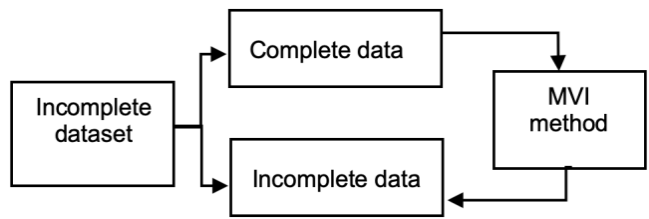

Fig. 2. Typical experimental setup for missing value imputation.

### B. Eligibility criteria and study selection

The citations and records obtained from the initial search were imported into Mendeley, a bibliography management software (version 1.16.3). Using the software, duplicate studies were identified and removed. The authors independently screened the titles and abstracts of the remaining articles. Full-text documents of potentially relevant articles were retrieved, added to Mendeley, and further assessed for eligibility. Inclusion and exclusion criteria for full-text eligibility were defined prior to the literature search and aligned with the identification process. The authors carefully evaluated the full-text articles for final inclusion in the review.

To be included, articles had to meet the following criteria: (1) utilization of at least one ML technique for missing value imputation, (2) reporting of quantitative outcomes, (3) original research papers, and (4) written in English. Retrospective studies, case reports, ongoing studies, surveys or reviews, simulation studies, protocol descriptions, qualitative studies, and studies published before 2014 were excluded to focus on recent research developments. The study identification and selection process were reported following the PRISMA flow diagram.

### C. Data extraction

This review had a specific focus on summarizing the use of ML techniques for modeling and evaluating the corresponding performance of these models. The selection of extraction items was based on previous literature discussing ML in the context of MVI [28], [24], [29] and was further refined through discussions among the authors. For the eligible articles included in this review, one author performed the data extraction, and the accuracy of the extracted information was validated by another author. Any discrepancies between the reviewers were resolved through discussion. Microsoft Excel was utilized for creating data extraction spreadsheets, which covered the following categories:

*1) Study characteristics*

This category included details such as the first author, publication year, data source, and the outcomes studied.

*2) Model performances*

This category captured information about the ML algorithms used, model descriptions, model validation, and model discrimination.

*3) Performance measures*

This category encompassed metrics like Mean Absolute Error (MAE), Root Mean Squared Error (RMSE) and accuracy, precision, recall, F1 score, as well as the methods employed to address the class imbalance problem.

*4) Quality assessment*

This category involved evaluating the quality of the included studies.

*D. Literature Search Outcomes*

This review utilized the well-known PRISMA strategy [26], depicted in Figure 3. By searching "missing value imputation" as the keyword on Google Scholar, from the search result, the first n = 500 articles were initially identified. all retrieved records were imported into Mendeley bibliography management software [27], which identified 14 duplicates and 3 non-English articles. The titles and abstracts of 483 unique papers were screened after removing duplicates. In the first screening round, review, database, or letter-type articles (81) were excluded. Only articles published between 2013 and 2023 were considered, resulting in n = 192 articles (as shown in Figure 3). We screened the abstracts of the remaining results according to our inclusion and exclusion criteria. Subsequently, a thorough examination was conducted on the selected articles, leading to the exclusion of 15 articles with irrelevant research objectives and 9 articles that did not match the required methods. This second round of exclusion yielded n = 168 articles. Finally, 49 articles were further excluded as they did not explicitly present MVI methods or relevant strategies. Through the subsequent stages of identification, screening, and eligibility testing, After the remaining 119 resulting citations in full-text form were assessed for full-text eligibility for the purpose of this review.

## III. Machine Learning Methods

Artificial Intelligence (AI), especially Machine Learning (ML), has experienced significant growth in recent years, revolutionizing data analysis and computing, enabling intelligent applications to function effectively. ML empowers systems to learn and improve from experience autonomously, without specific programming, making it a cornerstone of the fourth industrial revolution's latest technologies. Its application extends to various digital domains, including the Internet of Things, cybersecurity, mobile data, business, social media, and health data [30]. A ML model is a mathematical or computational representation of a system that can learn patterns from data and make predictions or decisions without being explicitly programmed. ML algorithms can be broadly classified into four primary types: supervised, semi-supervised, unsupervised, and reinforcement learning. Supervised learning uses labeled data to learn mapping from inputs to outputs, unsupervised learning finds patterns in unlabeled data, and reinforcement learning involves learning through interactions with an environment to maximize rewards.

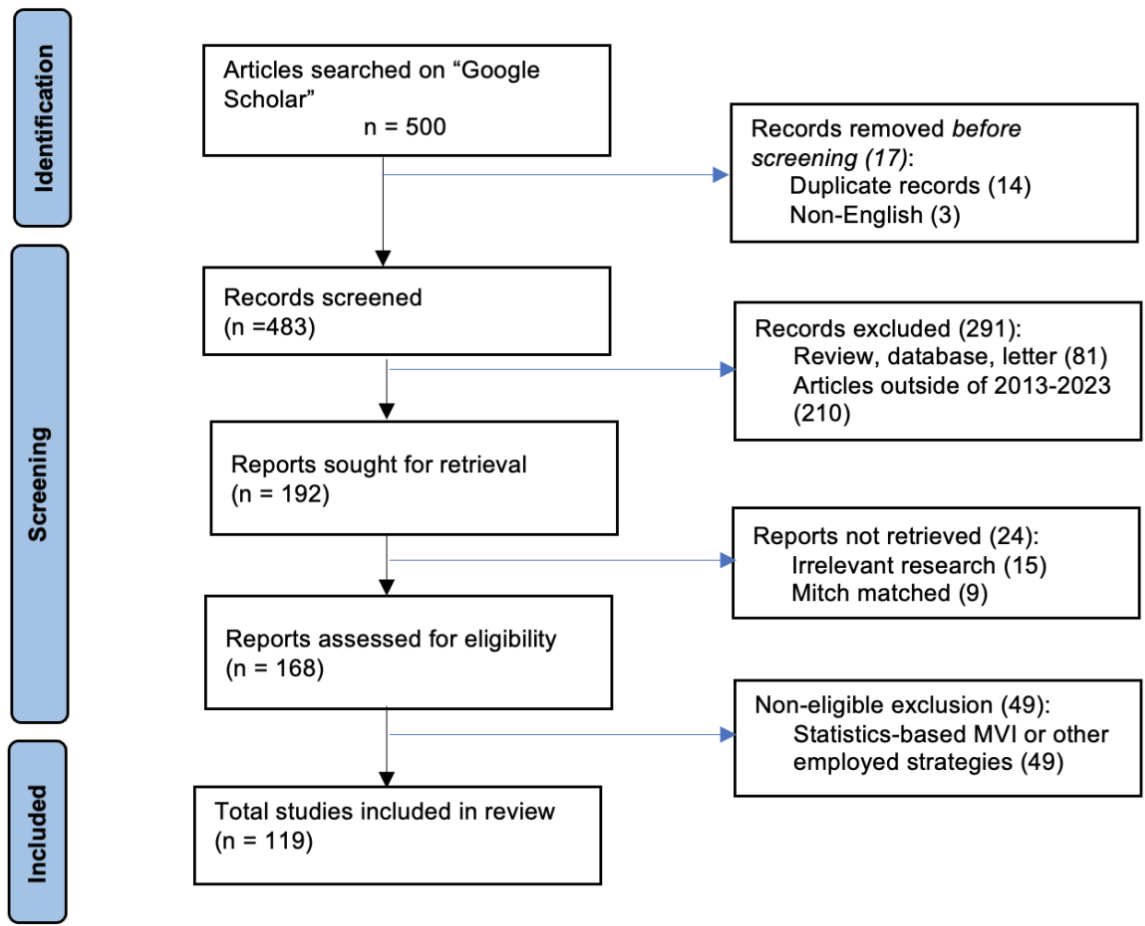

Fig. 3. PRISMA flow diagram.

Supervised learning, also known as the task-driven approach, typically involves the ML model learning a function that maps input to output based on sample input-output pairs [31]. In supervised learning, the ML model is trained on a labeled dataset, where the input data is associated with corresponding output labels. The model learns to map inputs to the correct outputs during the training process. The main goal of supervised learning is to make predictions on new data based on the patterns learned from the training dataset. The most common supervised tasks are classification and regression. The former task involves dividing the data into classes, while the latter regression task fits the data.

Unsupervised learning, a data-driven process, explores unlabeled datasets without the need for intervention. In unsupervised learning, the model is trained on an unlabelled dataset, meaning there are no corresponding output labels. The model learns to find patterns and structures in the data without any explicit guidance. Clustering and dimensionality reduction are common unsupervised learning tasks. Clustering involves grouping similar data points together, while dimensionality reduction aims to reduce the complexity of the data by extracting its essential features. [31]. Other popular unsupervised learning assignments include density estimation, feature learning, finding association rules, anomaly detection, etc.

Semi-supervised learning can be described as a hybridization of the supervised and unsupervised techniques mentioned above, as it works on both labeled and unlabeled data. Unlike supervised learning that relies solely on labeled data, and unsupervised learning that uses only unlabeled data, semi-supervised learning leverages both types of data to improve the model's performance. In real-world scenarios, obtaining labeled data can be costly and time-consuming, while unlabeled data is often more abundant and easier to collect. Semi-supervised learning bridges this gap by using the unlabeled data to augment the learning process and enhance the model's generalization capabilities. Semi-supervised learning has been successfully applied in various domains, including machine translation, natural language processing, computer vision, speech recognition, data labeling, and anomaly detection. [31], [32].

Reinforcement learning is a type of ML where an agent learns to make decisions by interacting with an environment. Reinforcement learning enables software tools and machines to automatically assess the optimal performance within a specific context, aiming to increase productivity. The agent takes actions in the environment and receives feedback in the form of rewards or penalties based on the consequences of its actions. The goal of reinforcement learning is to maximize the cumulative reward over time. It serves as a powerful tool for training AI models capable of supporting increased automation or optimizing the operational effectiveness of sophisticated methods, such as robotics, autonomous driving tasks, manufacturing, and supply chain logistics. However, it may not be the best approach for solving fundamental or straightforward problems [29], [31], [32].

The discussions lead to the categorization of ML models into five distinct groups: Classification, Regression, Clustering, Association rule learning, and Reinforcement learning. The first category, classification, represents a fundamental learning method used to address predictive modeling problems. In classification, the goal is to predict a class label or category for a given input example, based on the patterns learned from the training data [30]. This type of model is particularly useful when the output is categorical or discrete in nature, such as identifying whether an email is spam or not, or whether an image contains a specific object. Within the classification category, there are eleven different strategies or algorithms employed for building classification models. These strategies encompass various approaches to learning and prediction. Here are the details of the specific strategies:

1. Naive Bayes (NB): This algorithm is based on Bayes' theorem and assumes that features are conditionally independent given the class label. It's commonly used for text classification and spam filtering [33].

2. Decision Trees (DT): Decision trees partition the input space into regions based on feature values and make decisions by traversing the tree's branches. They are interpretable and can handle both categorical and numerical data [34].

3. Support Vector Machines (SVM): SVM aims to find a hyperplane that best separates different classes in the feature space. It's particularly effective for binary classification tasks and can handle complex decision boundaries [35].

4. Stochastic Gradient Descent (SGD): This optimization technique is used in conjunction with various classification algorithms to find the optimal weights for the model. It's commonly used for large-scale machine learning problems [36].

5. LGBM (LightGBM): LGBM is a gradient boosting framework that focuses on speed and efficiency. It builds trees leaf-wise instead of level-wise, leading to faster training times [37].

6. XGB (XGBoost): Similar to LGBM, XGBoost is another popular gradient boosting algorithm known for its high performance and regularization techniques [38].

7. Rule-based Classifiers (RbC): Rule-based classifiers generate a set of rules that can be used to classify instances. They are interpretable and can capture complex decision logic [39].

8. K-Nearest Neighbors (KNN): KNN assigns a class label to a data point based on the class labels of its k nearest neighbors. It's a simple and intuitive classification approach [40].

9. Random Forest (RF): Random Forest combines multiple decision trees to improve the model's performance and reduce overfitting. It's robust and can handle large datasets [41].

10. Logistic Regression (LR): Despite its name, logistic

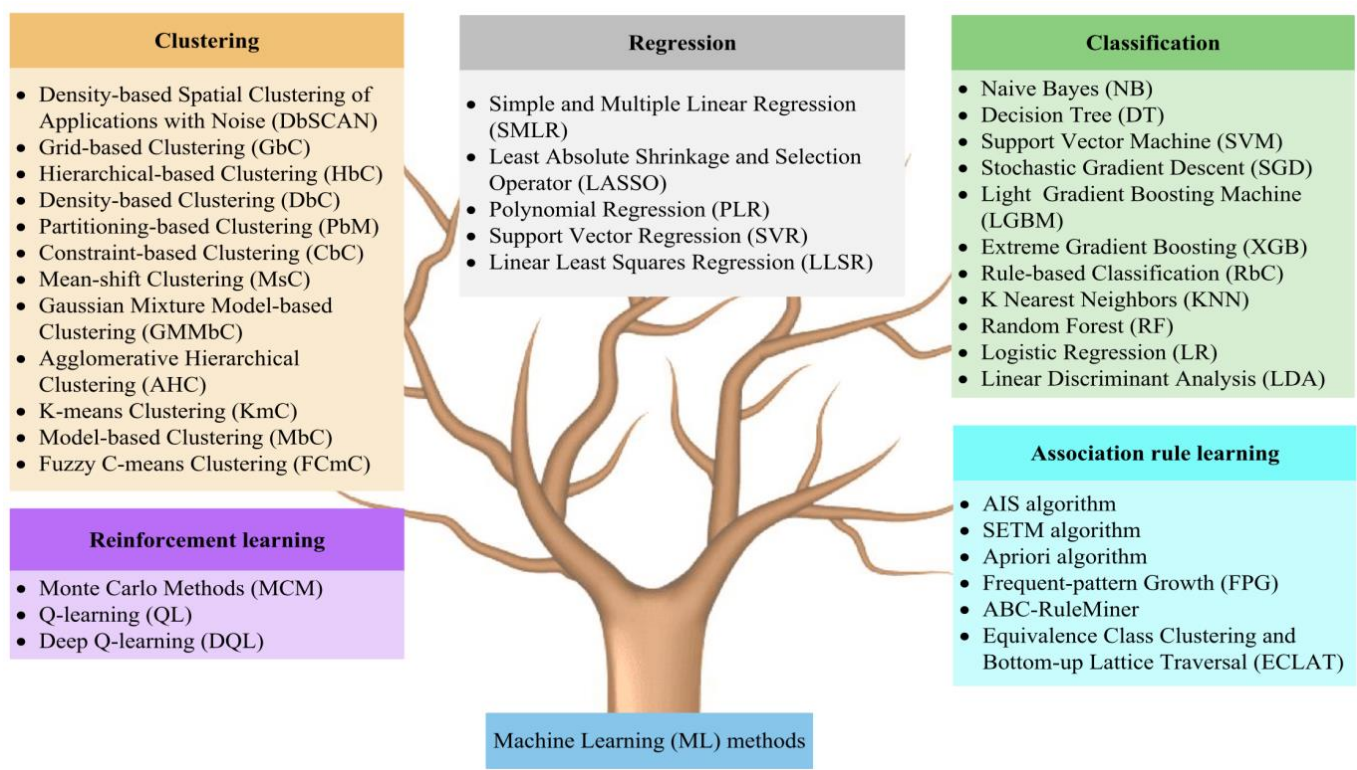

Fig. 4. ML models representation by grouping them based on their similarity [22].

regression is used for binary classification. It models the probability of the input belonging to a particular class [42].

11. Linear Discriminant Analysis (LDA): LDA is a dimensionality reduction technique that aims to find the linear combinations of features that best separate different classes [43].

Each of these strategies has its own strengths, weaknesses, and suitability for different types of data and problem domains. They collectively contribute to the arsenal of classification techniques used to predict class labels accurately based on input examples.

The second category, regression analysis, is a modeling approach used to predict a continuous outcome variable (denoted as $y$) based on the values of one or more predictor variables (denoted as $x$). Regression models are particularly useful when the relationship between the variables is quantitative, and the goal is to understand how changes in predictor variables impact the outcome variable. This category includes five distinct models, each with its own unique approach:

1. Sparse Multinomial Logistic Regression (SMLR): SMLR is a variant of logistic regression that handles multi-class classification tasks. It predicts the probability of an input belonging to each class and selects relevant features to improve model efficiency [30].

2. Least Absolute Shrinkage and Selection Operator (LASSO): LASSO is a linear regression technique that introduces regularization by adding a penalty term to the regression equation. It encourages feature selection by shrinking coefficients towards zero, effectively excluding less relevant predictors [44].

3. Polynomial Linear Regression (PLR): PLR extends regression by considering polynomial terms of predictor variables. It can capture nonlinear relationships between predictors and the outcome by introducing higher-order terms [45].

4. Support Vector Regression SVR: SVR is an extension of the support vector machine algorithm to regression tasks. It aims to find a regression line that has a maximum margin from the data points while allowing for a tolerance margin for errors [46].

5. Linear Least Squares Regression (LLSR): LLSR is the classical linear regression method that aims to minimize the sum of squared differences between predicted and actual outcomes. It finds the best-fitting line through the data points by adjusting the intercept and slope [47].

These regression models play a crucial role in predicting continuous values, such as predicting housing prices based on features like square footage, number of bedrooms, and location. Each model offers different advantages and is suitable for various types of data and modeling goals, allowing analysts to choose the most appropriate approach for their specific regression analysis tasks.

The third category, clustering, represents an unsupervised machine learning technique used to identify and group related data points within large datasets. Unlike supervised learning, clustering doesn't rely on specific outcome labels; instead, it focuses on finding patterns and structures within the data. This category encompasses twelve distinct approaches, each designed to address clustering tasks:

1. Density-Based Spatial Clustering of Applications with Noise (DbSCAN): DbSCAN identifies clusters based on the density of data points. It can find clusters of varying shapes and sizes, while also capturing noise and outliers [48].

2. Graph-based Clustering (GbC): GbC utilizes graph theory to identify clusters as connected components within a graph. It captures relationships between data points and their connections [49].

3. Hierarchical-based Clustering (HbC): HbC builds a hierarchy of clusters, organizing data points into a tree-like structure. It offers insights into various levels of clustering granularity [50].

4. Density-based Clustering (DbC): Similar to DbSCAN, DbC identifies clusters based on data point densities. It adapts the neighborhood size dynamically to capture different densities [51].

5. Partition-based Method (PbM): PbM divides the dataset into non-overlapping partitions, which are then refined into clusters. It's effective for handling large datasets and can capture clusters of different shapes [50].

6. Center-based Clustering (CbC): CbC identifies clusters around central data points. It often uses centroids or medoids to represent cluster centers [52].

7. Model-based Subspace Clustering (MsC): MsC is suitable for high-dimensional data by identifying clusters in subspaces. It considers the unique characteristics of data subspaces [53].

8. Gaussian Mixture Model-based Clustering (GMMbC): GMMbC assumes that data points are generated from multiple Gaussian distributions. It captures complex cluster shapes and can handle overlapping clusters [54].

9. Agglomerative Hierarchical Clustering (AHC): AHC starts with individual data points as clusters and progressively merges them based on similarity. It results in a hierarchy of clusters [55].

10. K-Means Clustering (KmC): KmC partitions the dataset into a predefined number of clusters by iteratively updating cluster centroids. It's a widely used and simple clustering method [56].

11. Model-based Clustering (MbC): MbC fits statistical models to data points within clusters. It can capture clusters of different shapes and sizes by using various probabilistic models [57].

12. Fuzzy C-means Clustering (FCmC): FCmC assigns data points to clusters with varying degrees of membership. It's suitable for cases where data points may belong to multiple clusters to a certain extent [58].

Each of these clustering approaches has its strengths and limitations, making them suitable for different types of data and clustering scenarios. They contribute to the task of identifying patterns and groupings within data without the need for labeled outcomes, providing valuable insights into the underlying structures of complex datasets.

The fourth classification, association rule learning, represents a rule-based machine learning approach aimed at uncovering intriguing relationships in extensive datasets. This method focuses on extracting "IF-THEN" statements that highlight connections between variables. This category encompasses six distinct models, each tailored to identifying and analysing such relationships:

1. Adaptive Intelligent System (AIS): AIS utilizes adaptive techniques to extract association rules from data, considering both frequent and rare itemsets. It adapts to changes in the dataset and its characteristics [59].

2. Sequential Ensemble Topical Model (SETM): SETM employs sequential ensemble learning to mine associations in sequences of data. It's particularly suitable for time-series and sequence data [60].

3. Apriori: Apriori is a classic algorithm that identifies frequent itemsets in the dataset and extracts association rules based on these frequent itemsets. It uses a bottom-up approach to generate rules [61].

4. Frequent Pattern Growth (FPG): FPG, like Apriori, identifies frequent patterns in the dataset, but it uses a more efficient pattern growth approach to mine associations. It avoids generating candidate itemsets [62].

5. Artificial Bee Colony Rule Miner (ABCRuleMiner): ABCRuleMiner is inspired by the foraging behavior of bees and uses artificial bee colony optimization to mine association rules. It explores the solution space to find relevant rules [63].

6. Equivalence Class Transformation (ECLAT): ECLAT efficiently identifies frequent itemsets using vertical data representation. It's suitable for high-dimensional datasets and can quickly discover associations [64].

Each of these association rule learning models offers a unique approach to discovering meaningful relationships within data. By identifying patterns and connections in "IF-THEN" statements, these models provide insights into the interactions between variables in large datasets, contributing to the understanding of underlying trends and correlations.

The fifth category, reinforcement learning, represents a machine learning technique that enables an agent to learn through a process of trial and error within an interactive environment. The agent learns by receiving feedback from its actions and experiences, adjusting its strategies to maximize rewards over time [30], [65]. This category encompasses three distinct methods:

1. Model-Based Control Method (MCM): MCM focuses on building a model of the environment and uses it to plan actions that maximize rewards. It combines model-based planning and control for effective decision-making [66].

2. Q-Learning (QL): QL is a widely used reinforcement learning algorithm that learns action-value functions through exploration and exploitation. It uses a table to store Q-values that represent the expected rewards for taking specific actions in certain states [67].

3. Deep Q-Network Learning (DQL): DQL extends Q-learning by utilizing deep neural networks to approximate Q-values for high-dimensional state and action spaces. It's particularly effective for complex environments and tasks [68].

Reinforcement learning is especially suited for scenarios where an agent interacts with an environment to learn optimal policies by receiving feedback on its actions. It's commonly applied in tasks such as robotics, game playing, autonomous navigation, and resource management. The methods within this category allow the agent to make informed decisions based on the experiences it gains over time, contributing to more efficient and intelligent decision-making processes.

## IV. MVI METHODS

In the context of MVI, a variety of methods have been explored in the literature [23], [22], [69], each offering different strategies for dealing with missing data. These methods can be broadly classified into two main categories: statistical techniques and Machine Learning (ML)-based techniques. The choice between statistical and ML-based techniques depends on factors such as the nature of the dataset, the extent of missingness, and the desired level of accuracy.

### A. Statistical Techniques

Statistical methods for MVI utilize mathematical and probabilistic approaches to estimate missing values based on the available data [70], [71], [72]. These methods often rely on summary statistics, relationships between variables, and probability distributions. Some common statistical techniques include:

**Mean or Median Imputation**: Replace missing values with the mean or median of the available data for that variable.

**Regression Imputation:** Use regression models to predict missing values based on relationships with other variables.

**Hot Deck Imputation:** Match each missing value with a similar observed value from the same dataset.

**Expectation-Maximization (EM) Algorithm:** Iterative approach to estimate missing values while considering their dependencies [70].

Statistical methods are often simpler and computationally efficient, making them suitable for cases with limited missing data. ML-based techniques, on the other hand, can capture complex relationships and are advantageous when dealing with more extensive missing data or datasets with intricate patterns. We focus on ML-based MVIs in this article.

### B. ML-based Techniques

Use ML-based MVI techniques leverage algorithms from the field of machine learning to predict missing values based on patterns and relationships in the data. These methods often provide more accurate imputations by considering complex interactions between variables. Some ML-based techniques include: KNN, DT, RF, Neural Networks, Matrix Factorization, Generative Adversarial Networks (GANs), etc. The selection of an appropriate MVI technique depends on the specific characteristics of the dataset and the goals of the analysis.

ML-based techniques include:

- **K-Nearest Neighbors (KNN):** Predict missing values based on the values of the nearest neighbors in the dataset.
- **Decision Trees:** Construct decision trees to predict missing values by traversing through the tree nodes.
- **Random Forests:** Ensemble of decision trees used to impute missing values.

- **Neural Networks:** Utilize neural networks to learn and predict missing values through hidden layers.
- **Matrix Factorization:** Decompose the data matrix into latent factors to predict missing values.
- **Generative Adversarial Networks (GANs):** Utilize GANs to generate missing values that align with the distribution of the data.

K-Nearest Neighbors MVI (KMVI) imputes missing values by employing the K-nearest neighbors approach. This method is suitable for scenarios where the Euclidean distance might not be the most effective distance metric. It draws missing values from the posterior predictive distribution, capturing the uncertainty associated with the imputed values [73]. However, KMVI's convergence can be slow, and it may not be the ideal choice when confidence intervals of estimated parameters are of primary importance. Improved K-Nearest Neighbors MVI (IKNNMVI) [74] imputes missing values by averaging nearest neighbors' values using the Euclidean distance metric. This approach takes into account correlations between different dimensions, offering a more comprehensive insight into the data structure. By focusing on the gray distance between the missing datum and its nearest neighbors, IKNNMVI reduces time complexity while minimizing errors associated with traditional Euclidean distance methods. The method aims to select the nearest neighbor instance with the same class label, contributing to improved time efficiency and enhanced accuracy in imputations.

Gray K-Nearest Neighbors MVI (GKNNMVI) [75] predicts missing values utilizing the gray distance-based K-nearest neighbors approach, which outperforms traditional Euclidean distance calculations. The gray coefficient guides the selection of valid attribute values derived from nearest neighbors, facilitating accurate imputations. GKNNMVI incorporates a normalized measure function and integrates mutual information-based feature relevance during the estimation of missing values, enhancing its ability to handle complex data patterns. Selective K-Nearest Neighbors MVI (SKNNMVI) [76] employs a weighted K-nearest neighbors approach that takes into account the correlations between different dimensions. By considering these correlations, SKNNMVI effectively addresses missing data imputation tasks. It fills missing values by calculating the mean of the nearest neighbor (NN) genes in the complete dataset [77]. The Euclidean Distance serves as the distance metric to determine the NN genes. This approach enhances accuracy and computational complexity compared to conventional K-nearest neighbors-based and other maximum likelihood estimation-based methods.

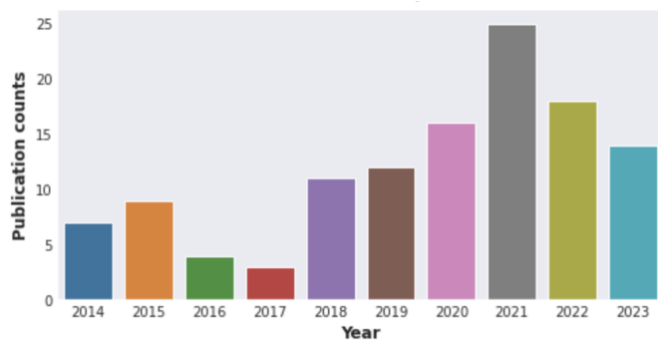

Fig. 5. Distribution of the published articles over the years considered in the literature review.

Weighted K-Nearest Neighbors MVI (WKNNMVI) [78] predicts missing values by utilizing the posterior predictive distribution. This approach is particularly useful for handling complex data distributions where traditional methods may fall short. It employs a weighted K-nearest neighbors strategy, assigning weight based on the correlation between a missing dimension and available data values from other fields. However, the performance of WKNNMVI depends on data quality and is sensitive to data scaling, which may lead to higher computational demands.

K-Means Clustering MVI (KMCMVI) [79] iteratively segments the dataset into K distinct clusters. Missing values are imputed based on the centroids of these clusters, effectively assigning data points to groups. The technique seeks to partition the dataset into K distinct clusters, optimizing the sum of squared distances between data points and the centroid of each cluster. It performs well for more or less linearly separable clusters but may struggle with global clusters. Fuzzy k-Means MVI (FKMMVI) [80] employs likelihood-based data assignment to clusters, iterating until the likelihood is maximized. It excels in handling overlapping clusters and is particularly suitable for tasks involving complex patterns. This method is employed to explore the structure of patterns when clusters overlap, aiming to capture the intricate nature of data. It minimizes the risk of getting trapped in local minima and proves advantageous when dealing with clusters that lack distinct separation. Similar to FKMMVI, Fuzzy C-Means MVI (FCMMVI) [81] assigns data to clusters based on likelihood, focusing on handling uncertainty in data clustering. This approach enables data points to belong to multiple clusters with varying likelihood, offering insights into the underlying patterns. The assignment process iterates to maximize the likelihood and achieve convergence. This technique is frequently used in pattern recognition tasks where data assignments are not confined to exclusive clusters.

Classification and Regression Trees MVI (CARTMVI) [82] utilizes decision trees to predict missing values, making it effective for categorizing or classifying data. It identifies predictors and cut points within the predictors to split the dataset into more homogeneous subsets. This technique can manage skewed numerical data distributions and multi-modal data, as well as categorical predictors with ordinal or non-ordinal structures. C4.5 Decision Trees MVI (C4.5MVI) [83] constructs decision trees and optimizes them through a pruning process. It is primarily designed for classification tasks and aims to strike a balance between accurate predictions and tree complexity. C4.5 employs a pruning mechanism where a subtree within the decision tree is replaced by a single decision node, consolidating all the decisions made by the subtree. However, this method can be unstable as minor changes in data might lead to significant variations in the optimal decision tree's structure. Additionally, C4.5MVI demands more memory resources for effective operation.

Random Forest MVI (RFMVI) [20] involves growing decision trees from randomly selected training samples and utilizing the most important features from these trees for imputation. This approach is effective for handling complex relationships within the data. Decision trees are constructed by selecting training samples randomly, with a focus on growing trees as much as possible without the need for pruning. In RFMVI, the importance of the top features in the trees outweighs that of the end nodes. However, it's important to note that RFMVI can exhibit bias when dealing with categorical variables. The use of multiple trees may also lead to slow algorithm performance and inefficiency in real-time predictions.

Artificial Neural Networks (ANNs) are frequently utilized for missing value imputation because of their capacity to discern intricate patterns and correlations within datasets [84]. The architecture of an ANN can be tailored to suit the specific characteristics of the dataset and the patterns of missing values. One approach, Multi-Layer Perceptron MVI, involves the creation of a multi-layer perceptron (MLP) model dedicated to predicting missing values for instances devoid of such data [85]. This technique harnesses feedforward neural networks to perform the imputation process. Initially, an MLP model is established using complete data instances, which is then employed to predict missing values for the relevant instances. However, in scenarios where multiple attributes contain missing data in high-dimensional problems, the need to construct numerous MLP models may adversely affect computational efficiency.

Different types of ANNs such as Feedforward Neural Networks, Recurrent Neural Networks (RNNs) [86], or Convolutional Neural Networks (CNNs) [87] can be employed depending on the nature of the data. Another MVI method utilizes an auto-encoder architecture to predict and impute missing values iteratively [88]. By harnessing the capabilities of end-to-end auto-encoders, this approach learns intricate data relationships and effectively predicts missing values. Neural Network based MVI proves particularly beneficial for addressing missing data patterns within datasets.

As ML-based imputation techniques [89], [90], [91] have shown great potential in recent years, researchers are increasingly exploring the use of deep learning approaches to tackle imputation tasks, especially in the domain of time-series data. Recurrent neural networks (RNNs) [86], [92], [93] have been the focus of much attention in this regard. For instance, one study introduced GRU-D, which incorporates a decay term into the GRU cell to help the model grasp the impact of temporal gaps between missing data points [86]. Another study used bidirectional RNNs to improve imputation accuracy [93].

Other innovative approaches have involved adding a weighted linear memory vector to RNNs [92], [94] inspired by ResNet, or training models only on observed values from incomplete datasets [86], [93]. Some studies have also used classification loss during training [92]. Moreover, transformer-based models are increasingly being explored in this field. For example, SAITS [95] proposed a diagonally-masked self-attention block for effective missing value generation, while MTSIT [96] introduced a method to train transformer architectures in an unsupervised manner.

In 2014, Goodfellow et al. [97] introduced Generative Adversarial Networks (GANs), which is a framework used to estimate generative models via an adversarial process. GANs consist of two components: a generator and a discriminator. The generator aims to deceive the discriminator by generating fake samples from a random "noise" vector. In contrast, the discriminator aims to differentiate between fake and real samples by identifying the probability that a sample originates from real datasets instead of the generator. However, traditional GANs are difficult to train due to issues such as mode collapse. Wasserstein GAN (WGAN) [98] provides an alternative training approach that enhances learning stability and mitigates mode collapse problems. Numerous studies have demonstrated that well-trained GANs can produce realistic images in computer vision applications [99], [100], [101], [102] and effectively complete facial images [103], [104], [105], [106]. While GANs are commonly used in image generation, they have been less explored in sequence generation tasks [107] with approaches like SeqGAN [108] and MaskGAN [109]. Most GAN-based sequence generation methods generate new samples from a random "noise" vector.

The task of data imputation necessitates imputed values to closely resemble the original incomplete data. Recent advancements in imputation methods utilizing generative models [110], [111], [112], [113], [114] have shown promise. Generative Adversarial Networks (GANs) [97] facilitate learning the distribution of original data and generating data with similar distributions, thereby enabling effective imputation. Yoon et al. [111] employed conditional GANs, using original data and masks as conditions. Luo et al. [113] introduced a recurrent network utilizing GRU-I, a simplified version of GRU-D, combined with GAN. Similarly, [112] utilized GRU-I but devised a recurrent autoencoder, representing a key distinction. In contrast to existing autoregressive approaches, [114] developed a non-autoregressive method and adopted a divide-and-conquer strategy to gradually fill missing values instead of all at once. Additionally, [110] implemented a conditional score-based diffusion model [115] for imputation, explicitly trained for this purpose and leveraging correlations between observed values.

In a two-stage approach, Luo et al. [113] utilized a GAN-based method for imputing missing values. Another notable method, GAIN (Generative Adversarial Imputation Network) [111], employs GANs with a hint vector to impute missing values.

Extreme Learning Machine MVI (ELMMVI) [116]: ELMMVI adopts a three-layer neural network architecture to predict missing values. Unlike some other neural network approaches, ELMMVI avoids multiple iterations and local minimization during the learning process. The model consists of input, hidden, and output layers, adhering to empirical risk minimization principles. It boasts improved computational efficiency due to the absence of numerous iterations and local optimization. ELMMVI's speed and promising performance make it stand out compared to existing neural network algorithms.

One commonly used imputation method is low-rank matrix/tensor completion. The Matrix Factorization (MF) method, explained in [117], aims to fill in the missing values of an incomplete matrix by factorizing it into low-rank matrices. A recent advancement in this area is the low-rank autoregressive tensor completion (LATC), introduced by [118], which considers temporal variation as a new regularization term for completing third-order tensors. Another MF-based MVI approach [119] that regularizes across factors to prevent their overinfluence and identifies underlying factors to model similarities across respondents and responses. Furthermore, Jia et al. [120] propose a novel Imputation Model for traffic Congestion data based on joint matrix factorization.

Association Rule MVI (ARMVI): ARMVI employs association rules to depict dependencies between data entries. It imputes missing values based on the inferred rules, making it suitable for datasets with intricate relationships [121]. This approach utilizes association rules to capture dependency links among data entries in a dataset, thus filling missing values according to those rules. Only database entries that perfectly match candidate patterns contribute to supporting those candidate patterns.

Genetic Algorithm MVI (GAMVI) [122]: GAMVI leverages genetic algorithms to generate optimal sets of missing values. It employs information gain as a fitness function to gauge the performance of individual imputation strategies. While GAMVI requires less data input about the problem, creating an effective objective function and selecting suitable operators can be intricate and computationally demanding. The method aims to create missing value sets that enhance the information gain, contributing to improved imputation outcomes.

Quantile Regression-based Imputation with Log-transformation for Censored Missing Values Imputation (QRILCMVI) is a technique used to impute missing values in datasets, particularly in scenarios where the missing data are left-censored, meaning that the true values are known to be below a certain threshold but the exact values are unknown [123].

The SVDMVI method, which stands for Singular Value Decomposition-based Missing Value Imputation, addresses missing values in datasets through an iterative process. It fills the missing values by computing a linear combination of the k most significant eigenvariables obtained through Singular Value Decomposition (SVD). This process iterates until a predetermined convergence threshold is reached, ensuring that the imputation is refined gradually. One of its key advantages lies in its capability to handle datasets with varying sizes and mixed data types efficiently [124].

Hierarchical Clustering MVI [125] constructs a hierarchy of clusters using either agglomerative or divisive methods. It identifies overlapping clusters and is capable of managing complex data structures. This method creates a hierarchy of clusters through a greedy approach, with two variations: agglomerative (bottom-up) and divisive (top-down). It doesn't require prior knowledge about the number of clusters and is relatively straightforward to implement, although it incurs a higher time complexity ($n2log(n)$) for n samples.

Self-Organizing Map MVI [126] employs self-organizing maps to segment data into clusters. This method is particularly well-suited for handling complex datasets with intricate relationships. It allows data points to be assigned to multiple clusters, much like FCMMVI and FKMMVI. However, its performance may be impacted by the predefined distance metric in feature space, causing issues with categorical data and mixed data types.

Rough Set MVI (RSTMVI) [127] employs rough set rule induction to extract association rules that describe patterns of missing data. This technique considers approximations and decision rules in order to infer these patterns. By applying rough set theory, RSTMVI provides a method for obtaining association rules that capture dependencies and relationships among missing data patterns. Unlike some other methods, RSTMVI does not require any preliminary or additional information, such as a probability distribution of the given data samples.

Support Vector Regression MVI (SVRMVI) [128] utilizes a Support Vector Regression (SVR) model to estimate parameters, with the goal of minimizing distances between the actual and predicted values. It proves effective for handling continuous variables and predicting missing values. SVRMVI directly estimates parameters by minimizing various distances, such as the maximum distance and Hausdorff distance, between the actual and predicted values. However, this method is best suited for interval data and relies on the presence of outliers and overlapping classes. It may also underperform when dealing with a higher number of features with a smaller number of training samples.

Semi-supervised learning (SSL) encompasses algorithms that leverage both labeled and unlabeled data [129], [130]. Among the favored SSL approaches is self-training, wherein an ensemble or teacher model generates pseudo-labels for unlabeled data. In the context of imputation tasks, missing values can be viewed as unlabeled data, while observed values serve as labeled data due to the absence of ground truth for missing values.

These methods offer diverse approaches to imputing missing values using various ML techniques. The choice of method depends on the characteristics of the data, the specific problem, and the desired outcomes.

## V. MVI Evaluation

Missing data is a common challenge in various domains, affecting the quality and reliability of data analysis. Missing Value Imputation (MVI) techniques play a crucial role in addressing this issue by estimating or filling in missing values based on existing data patterns.

MVI evaluation refers to the process of assessing the performance and effectiveness of Missing Value Imputation (MVI) techniques. Evaluating the performance of MVI techniques is essential to ensure accurate and reliable imputations. This section presents a comprehensive overview of MVI evaluation methods and their significance in enhancing data analysis outcomes. Here are the key aspects of MVI evaluation:

1) *Performance Metrics*
   To evaluate MVI techniques, various performance metrics are used to measure the quality of the imputed values. Common metrics include Mean Absolute Error (MAE), Root Mean Squared Error (RMSE), Mean Squared Error (MSE), and correlation coefficients. These metrics help quantify the difference between the imputed values and the actual values, providing insights into the accuracy of the imputation.

2) *Cross-Validation*
   Cross-validation is a widely used technique to assess the generalization capability of MVI methods. It involves splitting the dataset into training and testing sets multiple times, training the MVI model on the training set, and then evaluating its performance on the testing set. This helps determine how well the MVI technique performs on unseen data and avoids overfitting.

3) *Comparative Analysis*
   MVI evaluation often involves comparing the performance of multiple imputation techniques. Researchers may evaluate various methods using the same dataset and measure their respective performance metrics. This allows for a direct comparison of how well different techniques handle missing data and which one provides more accurate imputations.

4) *Bias and Variance*
   Evaluating the bias and variance introduced by MVI techniques is essential. A good MVI technique should minimize bias, which is the difference between the imputed and actual values, and also maintain low variance to ensure stability across different samples.

5) *Impact on Downstream Tasks*
   MVI evaluation extends to examining how imputed values enhance the performance of downstream analysis tasks such as predictive modeling, classification, and clustering. This assessment validates the practical utility of imputation methods.

6) *Sensitivity Analysis*
   By introducing varying proportions of missing data, sensitivity analysis evaluates the robustness of MVI techniques to different levels of missingness. This analysis helps understand the strengths and limitations of methods in diverse scenarios.

7) *Real-world Application*
   Evaluating MVI techniques in real-world applications is crucial to validate their effectiveness. Applying MVI methods to practical datasets from various domains helps assess their performance in solving real data challenges.

8) *Comparison with Ground Truth*
   In some cases, when the true values of missing data are known, researchers can directly compare the imputed values with the actual values to evaluate the accuracy of MVI methods.

Effective evaluation of MVI techniques ensures accurate and reliable imputations, leading to improved data analysis outcomes. MVI evaluation involves using performance metrics, cross-validation, comparative analysis, and other techniques to assess the accuracy, bias, variance, and practical effectiveness of Missing Value Imputation methods. The goal is to identify the most suitable technique for imputing missing data in a given context and ensure reliable and accurate downstream analysis.

## VI. DISCUSSION

In this section, we will analyze the different methods discussed in the preceding sections. We will identify the scientific problems associated with the experimental framework and suggest them as potential future research objectives to advance MVI methods. Furthermore, we will explore how MVI techniques can enhance the outcomes of ML models when used with ML classifiers. We will also provide insights into recent research trends related to these methods gathered from articles published over the past decade. Finally, we will recommend further areas of exploration based on our findings.

The pie chart (Figure 6) illustrates the distribution of publications across different machine learning methods. Among the categories depicted, "NN" (Neural Networks) stands out as the most frequently utilized method, constituting approximately 24.4% of the total publications. K-Nearest Neighbors (KNN) and Generative Adversarial Networks (GAN) also make significant contributions, representing approximately 16.5% and 15.0% of the publications, respectively. Decision Trees (DT) and Random Forests (RF) have relatively smaller shares, with each accounting for around 3.9% and 4.7% of the total publications, respectively. Conversely, "MF" (Matrix Factorization) appears to have the lowest representation, with only about 3.1% of the publications attributed to this method. Overall, the pie chart provides a clear visual representation of the distribution of publication counts across different machine learning methods, highlighting the prominence of Neural Networks and the diversity of approaches captured under the "Others" category.

Neural Network (NN) based models play a crucial role in missing value imputation due to their ability to capture complex patterns and relationships within data. There are several reasons why NN-based models are important for missing value imputation such as non-linearity, automatic feature learning, adaptability to different data types and scalability. Neural networks are capable of modeling non-linear relationships between variables. This feature helps them to capture intricate patterns in data that may go unnoticed by linear imputation methods. One of the major advantages of NN-based models is that they can learn relevant features from the data automatically. They can handle different types of data, including numerical, categorical, and textual data, making them a versatile tool for imputing missing values in diverse datasets. NN-based models can learn end-to-end mappings from input to output, allowing them to directly impute missing values without the need for intermediate steps or pre-processing. Additionally, neural networks can scale to large datasets and complex models, making them well-suited for imputing missing values in big data scenarios. These features make NN-based models more efficient and accurate in handling missing data.

The KNNMVI method is widely used because of its ability to adapt to new data, straightforward algorithm, and fast computation times. Researchers have developed innovative techniques rooted in statistical modeling, machine learning, and, more recently, deep learning. Regression-based imputation methods harness the power of linear and nonlinear models. Decision trees and ensemble methods have also gained prominence for their ability to handle missing values effectively.

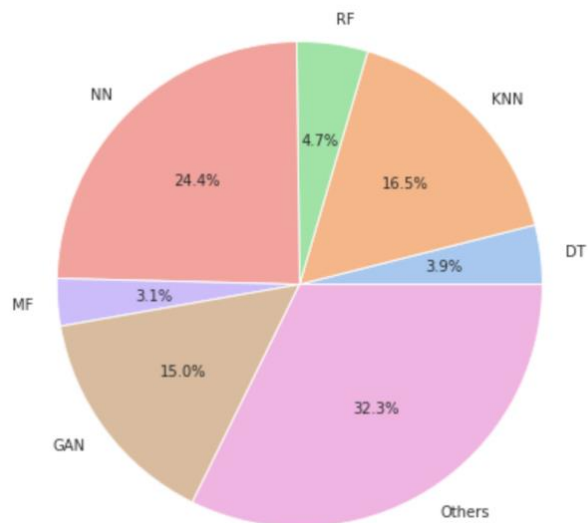

Fig. 6. A pie chart displaying the distribution of machine learning methods in percentage of publications attributed to that method.

Time-series data often exhibit non-linear relationships and complex patterns that classical statistical imputation methods may not capture. While machine learning and deep learning techniques offer improved flexibility, they are not without challenges and limitations. Understanding these challenges is crucial for researchers and practitioners seeking to apply imputation methods effectively. The effectiveness of imputation methods depends on various factors, including the type of missing data, data distribution, and the presence of temporal dependencies. In high-dimensional time-series data, where multiple variables are recorded over time, the complexity of imputation grows significantly. Traditional imputation methods may struggle to handle the curse of dimensionality, leading to increased computational requirements and potential overfitting. Effective feature selection and dimensionality reduction techniques are often needed to mitigate this challenge.

Deep learning has brought in a new era of data imputation with RNNs, LSTM networks, and TCNs showing remarkable capabilities in modeling complex temporal dependencies and adapting to irregularly sampled data. Deep learning-based imputation methods may demand substantial computational resources, especially for large-scale data. Training complex neural networks can be time-consuming and require access to high-performance computing infrastructure. This limitation can hinder the practicality of deploying deep learning imputation techniques.

Generative Adversarial Networks (GANs) have emerged as promising models for missing value imputation due to their ability to learn the underlying data distribution and generate realistic data samples. By learning the data distribution, GANs can generate plausible values for missing data points based on the observed data, leading to more accurate imputation results. GANs have the capacity to capture complex patterns and dependencies within the data, making them well-suited for imputing missing values in datasets with intricate relationships between variables. The generator in a GAN is capable of learning to generate data samples that preserve the statistical properties and correlations present in the observed data. GANs can learn from unlabeled data by leveraging the information contained in the observed data to generate realistic imputations for missing values. As research in GAN architectures and training techniques continues to advance, GAN-based imputation methods are expected to become increasingly effective and widely adopted in practical applications across various domains.

In the case of supervised techniques, such as KNNMVI, RFMVI, and CARTMVI, the complete dataset serves as the training set, and the dataset with missing values functions as the testing set. On the contrary, unsupervised clustering methods like KMCMVI and FCMMVI group related entities into coherent clusters. It is worth noting that each cluster's center is the average of all the entities within that cluster. In order to fill in missing values, the distance between incomplete data and recognized cluster centroids is calculated, and the values of the nearest centroid are used to fill in the gaps. However, it's important to note that such clustering methods can be time-consuming to reach convergence if the initial estimation significantly deviates from the actual solutions.

Evaluation plays a critical role in validating the performance of MVI methods and arriving at definitive conclusions. Evaluating the performance of imputation methods can be challenging due to the absence of ground truth for imputed values. It often relies on comparing imputed values to observed data, which may not fully capture the true underlying values. Additionally, imputation evaluation metrics may not account for the downstream impact of imputed values on subsequent analyses. A few research publications that have adopted both direct and indirect evaluation methods are [14], [131], [132], [133] [134], [135].

In practice, the choice of imputation method depends on the specific characteristics of the data, the domain of application, and the desired outcomes. Researchers and practitioners must carefully assess the appropriateness of models for their particular applications. Specially, In sensitive applications like healthcare, finance, and social sciences, imputation methods may introduce ethical considerations. Imputed values can influence critical decisions and policies, raising concerns about fairness, transparency, and accountability. Ensuring responsible and ethical use of imputation techniques is essential.

## VII. OBSERVATIONS AND RECOMMENDATIONS

The essential insights drawn from our meticulous examination of the chosen 119 publications, are presented in this section. Consequently, we offer various suggestions for the imputation of missing values, serving as future guidance for scholars at all levels, from beginners to experts, in constructing effective decision-making frameworks for noisy datasets. Our noteworthy observations and recommendations are outlined below:

• The need to employ Missing Value Imputation (MVI) processes hinges on the significance of the missing values. Novel classification or regression approaches exist that inherently handle data missingness without the need for MVI methods [136], [137]. Researchers can initiate pipeline development by obtaining approximate outcomes from these models, using both incomplete and complete datasets (processed by representative baseline MVI methods as suggested in this article). Such experiments can illuminate the significance of the missing values and help determine whether missingness imputation is essential.

• Our review analysis underscores that there is no definitive answer to the question of the "best" imputation method for missing values. The selection of MVI methods depends on several factors of interest. However, the insights gathered from the chosen papers, which incorporate various MVI methods to enhance the performance of ML models, indicate that the choice of MVI methods depends on various aspects. These aspects include the quality of the dataset, selection of feature attributes, outlier detection, normalization, application domains, available computational resources, time constraints for imputation, and the objective of preserving either means or inter-attribute correlations.

• The order in which attribute and/or sample selection is employed in relation to the MVI process can have an impact on the results of imputation [138], [139]. Conducting these operations before or after the MVI process can lead to cleaner complete datasets for learning or enhance the classifier's performance. While some articles have considered these effects [140], [141], [142]. Huang et al. [140] and Tsai and Chang [141] demonstrated that applying MVI method after sample selection yields better performance in downstream ML tasks. According to Huang et al. [140] and Liu et al. [143] the order in which attribute and/or sample selection is applied in relation to the MVI process can affect the imputation results. It has been revealed that attribute selection before the MVI process may not have a significant impact, while applying MVI before attribute selection can be advantageous. Additionally, Tran et al. [142] conducted experiments and confirmed that utilizing MVI before attribute and sample selection considerably enhances classification accuracy and reduces computational time.

• Data normalization or standardization plays a critical role in the imputation process. Failing to normalize or standardize the data in proper way may result in suboptimal imputation results, particularly for distance-based methods like K-Nearest Neighbors MVI. Standardizing the data before imputation is advisable, as lower attribute values lead to faster convergence, especially in real-time applications. A debate arises over whether to impute before or after standardization. Imputing first can affect the center and scale used in standardization. Conversely, standardizing first can impact the imputation procedure by influencing the center and scale employed.

• The presence of outliers can render imputed values unreliable, impacting the integrity of missing data imputation. Therefore, it is imperative to address outliers before conducting missingness imputation. Regression-type MVI methods are particularly sensitive to outliers and require mandatory outlier handling. Conversely, median value-based MVI methods can attenuate the influence of outliers, as they impute missing values based on the most frequent values, which are less likely to be outliers.

• In evaluating MVI methods, three common approaches emerge: direct assessment, specified metrics of classifiers for indirect evaluation, and consideration of Total Classification Time (TCT). A comprehensive understanding of MVI technique performance and the formulation of reliable imputation strategies necessitate the application of all three evaluation methods. Unfortunately, these approaches are seldom applied together in relevant studies over the past decade, indicating a shortcoming in current practices. Researchers should aim to integrate these evaluation methods in future experiments for a more holistic assessment.

## VIII. CONCLUSIONS

The process of Missing Value Imputation (MVI) for incomplete datasets holds pivotal significance within the domains of data mining, big data analysis, and ML-based decision-making pipelines. The integrity of mining outcomes and analytical results hinges upon the judicious imputation of missing data, making this a critical concern. In this study, 117 research papers were examined to explore various aspects of experimentation, including dataset characteristics, accuracy evaluation metrics, and imputation techniques such as algorithm categories, addressed missing mechanisms, and imputable data types. The results show that there is no one-size-fits-all solution to dealing with missing data challenges in real-world datasets. Different ML approaches work well depending on the characteristics of the dataset. KNN-based imputation is popular due to its simplicity and ease of implementation, while neural networks and deep learning-based methods are gaining increased attention. MVI plays a pivotal role in ensuring the integrity, reliability, and usefulness of time-series data across various applications and domains. The advent of deep learning and generative adversarial networks (GANs) has revolutionized the field by enabling the capture of intricate temporal dependencies and handling complex data patterns. As the field of missing value imputation continues to advance, researchers and practitioners are presented with exciting opportunities and challenges.